\documentclass{article}
\usepackage{ijcai24}

\usepackage{times}
\usepackage{soul}
\usepackage{url}
\usepackage[hidelinks]{hyperref}
\usepackage[utf8]{inputenc}
\usepackage[small]{caption}
\usepackage{graphicx}
\usepackage{amsmath}
\usepackage{amsthm}
\usepackage{booktabs}
\usepackage{algorithm}
\usepackage{algorithmic}
\usepackage[switch]{lineno}

\usepackage{mathtools} 
\usepackage{multirow} 
\usepackage{dsfont}
\usepackage{amssymb}
\usepackage{xcolor}
\usepackage{subcaption}
\usepackage{svg}

\title{By Fair Means or Foul: Quantifying Collusion in a Market Simulation with Deep Reinforcement Learning}

\author{
Michael Schlechtinger$^1$
\and
Damaris Kosack$^2$\and
Franz Krause$^1$\And
Heiko Paulheim$^1$\\
\affiliations
$^1$University of Mannheim - Chair of Data Science\\
$^2$University of Mannheim - Chair of Law\\
\emails
\{schlechtinger, damaris.kosack, franz.krause, heiko.paulheim\}@uni-mannheim.de
}

\begin{document}

\maketitle

\begin{abstract}
In the rapidly evolving landscape of eCommerce, Artificial Intelligence (AI) based pricing algorithms, particularly those utilizing Reinforcement Learning (RL), are becoming increasingly prevalent. This rise has led to an inextricable pricing situation with the potential for market collusion. Our research employs an experimental oligopoly model of repeated price competition, systematically varying the environment to cover scenarios from basic economic theory to subjective consumer demand preferences. We also introduce a novel demand framework that enables the implementation of various demand models, allowing for a weighted blending of different models.
In contrast to existing research in this domain, we aim to investigate the strategies and emerging pricing patterns developed by the agents, which may lead to a collusive outcome. Furthermore, we investigate a scenario where agents cannot observe their competitors’ prices. Finally, we provide a comprehensive legal analysis across all scenarios. 
Our findings indicate that RL-based AI agents converge to a collusive state characterized by the charging of supracompetitive prices, without necessarily requiring inter-agent communication. Implementing alternative RL algorithms, altering the number of agents or simulation settings, and restricting the scope of the agents’ observation space does not significantly impact the collusive market outcome behavior.
\end{abstract}

\section{Introduction}

The usage of pricing algorithm technologies has become ubiquitous within the realm of ECommerce platforms. In the last decade, these algorithms have shifted from static heuristics to AI-driven software that outperforms them in terms of average daily profits~\cite{kroppDynamicPricingProduct2019,qiaoDistributedDynamicPricing2024}. Studies have already shown that in certain markets it is unreasonable to conduct business at a profitable level lacking this technology, as algorithmic price setting frequencies create a significant competitive edge~\cite{assadAlgorithmicPricingCompetition2020}. 

Nevertheless, there is concern that these conditions harm competition and thus consumer welfare: The use of AI-based pricing algorithms often results in an inextricable pricing situation with a collusive market outcome. If the emergence of this collusive market outcome can be attributed to any form of concerted action, it gives rise to significant legal and ethical apprehensions.\footnote{For a more detailed introduction to Art. 101 Treaty on the Functioning of the European Union ("TFEU"), see Schlechtinger et al.,~\shortcite{schlechtingerWinningAnyCost2021}.} There is a necessity to explore the factors that contribute to the collusive market outcome of RL agents and shed light on the vulnerabilities and potential risks associated with their deployment in pricing algorithms. In pursuit of this goal, we endeavor to elucidate their underlying mechanisms and enhance the transparency and comprehensibility of coordinative AI-pricing strategies.

If no such coordination can be identified, the collusive market outcome generally must be considered legally neutral. While collusion may harm consumers, it is considered undesirable for innovation and economic growth from a welfare economic standpoint. The challenge from a doctrinal perspective lies in the difficulty of attributing legal responsibility directly to a market outcome~\cite{europeancommissionGuidelinesApplicabilityArticle2011}.

This evokes three major legal questions. First, does the agent's behavior constitute a minimum degree of coordination and thus violate the cartel prohibition under existing German and European competition law? Second, do the existing rules readily fit an algorithmic conduct or do we face a regulatory gap within the cartel prohibition? Third, should we expand cartel law to encompass algorithmic collusion?

\section{Related Work}
Drawing from economic and AI literature presents tendencies of AI-based algorithms to reach seemingly collusive outcomes. As such, Calvano et al. \shortcite{calvanoArtificialIntelligenceAlgorithmic2018} provide fundamentals for algorithmic pricing studies. While preceding research usually formulated a Cournot oligopoly model where two agents sell certain quantities of a good~\cite{waltmanQlearningAgentsCournot2008,kimbroughLearningColludeTacitly2009,siallaganAspirationBasedLearningCournot2013}, they transitioned towards a simplified oligopoly model based on a Bertrand price setting scenario. The agents used q-learning to “consistently learn to charge supracompetitive prices, without communicating with one another". The main contribution of this research is the illustration of a reward-punishment scheme among autonomous pricing agents. By manually precipitating exogenous price cuts for selected agents, the study shows that the algorithms are able to punish any behavior that deviates from a collusive state in order to gradually return back to it.

Other scholars investigated the issue by adding human participants~\cite{wernerAlgorithmicHumanCollusion2021}, by specifically analyzing sizes of discrete action spaces~\cite{kleinAutonomousAlgorithmicCollusion2021}, or by building custom, more elaborate scenarios~\cite{abadaArtificialIntelligenceCan2023}. Few researchers applied deep neural networks. The ones that did, were able to improve on Calvano et al.~\shortcite{calvanoArtificialIntelligenceAlgorithmic2018} by achieving a shorter learning time due to the usage of deep Q-Networks as well as reward averaging~\cite{hettichAlgorithmicCollusionInsights2021}. Others restricted algorithms to only memorize the periods when they do not exceed in terms of profits and ignored the ones when they outperform. Scholars, nevertheless, emphasize that “more efforts are needed in exploring other architectures of deep networks"~\cite{hanUnderstandingAlgorithmicCollusion2021}. Due to the characteristic learning behavior of RL-algorithms or AI, the quality of learning data significantly influences agents' propensity for collusion. To test this hypothesis, some scholars employed a simple upper confidence bound bandit algorithm to set a discrete number of prices~\cite{hansenAlgorithmicCollusionSupracompetitive2020}. Their outcome indicated that the prices are bound to the signal-to-noise ratio of their inputs, resulting in a supracompetitive state for less noisy input data and vice versa.

The current state of research is mainly simulation-based, with few scholars collecting empirical data. An investigation into Germany's retail gas market, conducted using a catalog of potential characteristics, identified widespread adoption of pricing algorithms since 2016. Consequently, sellers achieved margins above competitive levels. As their data indicates no initial effects, followed by an eventual convergence to high prices and margins, they infer that the algorithms were able to learn tacitly collusive strategies over time~\cite{assadAlgorithmicPricingCompetition2020}. Brown and MacKay~\shortcite{brownCompetitionPricingAlgorithms2021} tackle the issue from a different angle. The authors extract pricing data from five pharmacy firms with differing price changing frequencies~\cite{brownCompetitionPricingAlgorithms2021}. Musolff~\shortcite{musolffAlgorithmicPricingFacilitates2022} employs a dataset acquired from Amazon's buybox, an algorithmic pricing-heavy feature used by third-party sellers, to show that repricers have been able to avoid the competitive behavior by regular price raises. 

In essence, the major shortcomings of current research are the substantial deviation from realistic market models, the over-representation of tabular q-learning, a comprehensive legal analysis of the experiments, as well as the low density of empirical studies. Our work aims to bridge the gap between real-world empirical analyses and purely theoretic models by providing a simplified, but scalable market simulation. Thus, we propose a novel demand framework that enables the implementation of various demand models and facilitates integration between them, allowing for a weighted blending of different models. Within the framework, we investigate the behavior of a scalable amount of agents that rely on state-of-the-art deep reinforcement learning technologies (i.e., PPO, DQN). By understanding the underlying causes of collusive outcomes using an interdisciplinary approach, we contribute to the ongoing efforts of building AI systems that align with societal values and objectives.

\section{Experiment Design}

We consider an oligopoly setting at the core of our experiments. This fundamental stage game comprises $m \in \mathbb{N}$ consumers $Y=\{y_1,...,y_m\}$ and $n \in \mathbb{N}$ firms $x=\{x_1,...,x_n\}$ that simultaneously set the prices $P = \{p_1, ..., p_n\}$ so that $p_i\in [0,2]$ holds for all $i \in \{1, ..., n\}$. Accordingly, we define a general \textit{selling} or \textit{demand probability} as follows:

\begin{equation}
    d:=d(\Omega):\{1,...n\}\rightarrow[0,1]^n \;\; \text{where} \smashoperator[r]{\sum_{i \in \{1,...,n\}}} d_i=1
\end{equation}
The parameter $\Omega$ represents the buyers' background knowledge, allowing an implementation of a custom buying probability. For example, $\Omega$ might entail domain data such as the demand function of a market, leading to new demand probabilities. Given $p_\text{min} = \min_{1 \leq i \leq n} (p_i)$ and $P_{min} := \left\{ p \in P: p=p_{min} \right\}$, we can then define a Bertrand selling probability $d^b:\{1,...,n\}\rightarrow[0,1]^n$ via
\begin{equation}
d^b_i =
  \begin{cases}
    0, & p_i \neq p_\text{min}\\
    \frac{1}{|P_\text{min}|},  &p_i = p_\text{min}
  \end{cases},
\label{eq:bertrand-d}
\end{equation}
thus exclusively depending on the prices $P$.
According to Bertrand~\shortcite{bertrandTheorieRichessesRevue1883}, sellers will end up in a Nash equilibrium, represented by a price equal to the marginal costs (i.e., the competitive price) due to the buyers consistently purchasing the lowest-priced product. In a classic Bertrand oligopoly setting, the goods are characterized as perfect substitutes. However, this buying behavior is based on a theoretic construct as homogeneous goods may still be acquired from diverging sellers due to subjective consumer preferences in a realistic scenario. We aim to counteract this shortcoming from two sides. Legal research states that a relevant product market comprises all those products and services which are regarded as interchangeable or substitutable by the consumer, because of the products‘ characteristics, their prices, and their intended use~\cite{commissionregulationecno447/98NotificationsTimeLimits1998}. To combat this from an IT perspective, we used a selection strategy proposed by~\cite{zhongComparisonPerformanceDifferent2005}. While solely relying on prices, this approach allows us to create a simulation that counteracts the Bertrand model's main limitation: Its extremely punishing nature, which might complicate the learning process or force the agents into a collusive state. Furthermore, with the roulette wheel, we can model \textit{switching barriers} (i.e., expenses consumers feel they experience by switching from one alternative to another). Based on $\hat{p}_\text{max}$ as the maximum price achievable in a market scenario, this modification results in the buying behavior

\begin{equation}
d^r_i =
    \frac{\hat{p}_\text{max} - p_i}{\sum_{j\in\{1,...,n\}}(\hat{p}_\text{max} - p_j)}.
\label{eq:roulette-d}
\end{equation}

In order to bridge theory and empiricism, we introduce the factor $\mu\in [0,1]$. $\mu$ serves as a weight to gradually transition from one buying behavior to another. With this work, we combine the previously defined selection strategies $d^b$ in (\ref{eq:bertrand-d}) and $d^r$ in (\ref{eq:roulette-d})
with $\sum_{i=1}^n d^b_i = 1 = \sum_{i=1}^n d^r_i$ by means of
\begin{equation}
    \resizebox{.91\linewidth}{!}{$
        d^{\text{comb}, \mu} = \mu * d^b + (1 - \mu) * d^r  \;\; \text{with} \;\; \smashoperator[r]{ \sum_{i=1}^n} d^{\text{comb}, \mu}_i = 1.
    $}
\end{equation}

Due to this specific combination, $\mu$ acts as a bias.\footnote{We restrict $d^\text{comb}$ to $d_b$ and $d_r$ in this work. However this approach can be analogously extended to $k \in \mathbb{N}$ buying behaviours via biases $\mu_1,...,\mu_k$ if $\sum_{j=1}^k \mu_j = 1$.} If it is set to 1, the products are perfect substitutes, if it is set to 0, the consumers' buying behavior is regulated by the roulette selection (this means $d^{\text{comb}, 0} = d^r$ and $d^{\text{comb}, 1} = d^b$). With this in mind, we can switch from a theoretical Bertrand model to a more realistic setting, that involves subjective consumer preferences.

\begin{figure}[t]
  \centering
  \includegraphics[clip, trim=0.4cm 0.4cm 0.4cm .79cm, width=0.9\linewidth]{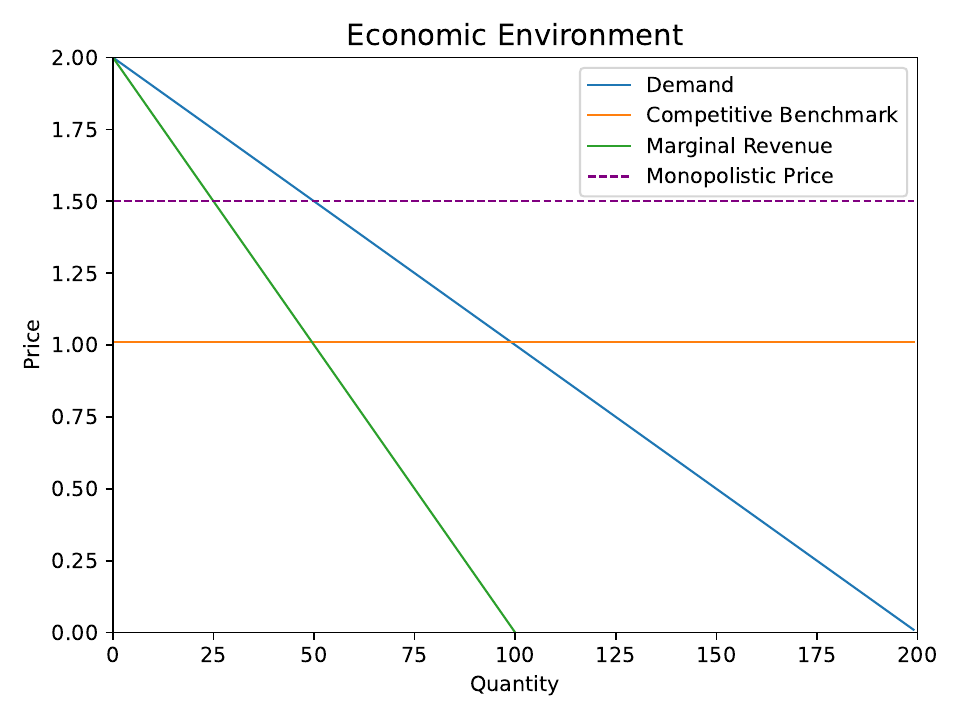}
  \caption{Economic settings within the environment}
  \label{fig:eco}
\end{figure}

With the parameters set, a seller can achieve a monopolistic price (MP) (i.e., the price that relates to the maximum revenue a monopolist can achieve in the market) of 1.50 (with a cumulative quantity of 50) and a competitive benchmark (CB) (i.e., the price one unit above the marginal costs) of 1.01 (with a cumulative quantity of 99) (cf. Figure \ref{fig:eco}). To create comparable results, we restrict the number of consumers to $m = 200$ and thus result in a maximum price $\hat{p}_\text{max} = 2$. We chose these specific model settings based on the demand function. We need to choose arbitrary confinements in order to satisfy the presented economic rules. The agents can set prices above and below these confinements.

\subsection{Deep Reinforcement Learning Algorithms}

While the implementation and analysis of deep RL algorithms are more complex, we benefit from conditions that more likely resemble state-of-the-art pricing algorithms, as the number of possible states and variables of a real-world market environment most likely overstrain basic reinforcement learning tables. By including multiple, heterogeneous RL technologies we aim to scrutinize the robustness of our results further. However, our selection is confined to model-free methods as these are generally more popular, quicker to implement, and more extensively developed and tested than model-based methods~\cite{achiamSpinningDeepReinforcement2018}. Our selection includes Deep Q-Networks (DQN)~\cite{mnihPlayingAtariDeep2013} and Proximal Policy Optimization (PPO)~\cite{schulmanProximalPolicyOptimization2017} with the intention of featuring an algorithm for both sides of the model-free taxonomy (q-learning and policy optimization). The algorithms also support the use of discrete action spaces, reducing the number of possible error sources.

\subsection{Markov Decision Process}
When an agent tries to maximize its interests from the economic environment, it must consider both the reward it receives after each action and the feedback from the environment. This can be simplified as a Markov Decision Process (MDP)~\cite{vanotterloReinforcementLearningMarkov2012}, more specifically, a  Partially Observable Markov Process (POMDP)~\cite{hausknechtDeepRecurrentQLearning2015} due to the multi-agent environment hiding sensitive information from the competitors. We choose this method, as the main issue of this paper can only be solved by various agents conducting subjective observations, due to which future game states will depend on more than just a single agent's current input. As MDPs are step-based and our previous illustrations were more general, we introduce time step $t$ to our established settings. We consider a sequential decision-making problem in which every agent (seller) $i \in \left\{ 1,...,n \right\}$ interacts with a stochastic Multi-Agent Reinforcement Learning (MARL) environment. Each agent at time step $t$ observes a state $s(t) = \big(p_i(t), ..., p_n(t), c \big) \in S$ where $p_i(t)$ is the price set by agent $i$ at time $t$, $c$ are the costs to purchase a good and $S$ is the global state space. For every time $t$, an agent $i$ takes an action $a_i(t) \in \mathcal{A}$ where $\mathcal{A}$ is the valid, discrete action space, and executes it in the environment to receive a reward $\epsilon_i (t)  \in \mathbb{R}$

\begin{equation}
  \epsilon_i(t) = m * \big[p_i(t) - c \big] * d_i(t),
\label{eq:profit}
\end{equation}

where $d(t)=\big( d_1(t),..,d_n(t) \big)$ is the previously defined selling or demand probability at time $t$.
Given the state $s(t)$ at time step t, the new state $s(t+1)$ is to be reached after carrying out the actions $A(t):=\big(a_1(t),...,a_n(t)\big) \in \mathcal{A}^n$ with the probabilities $\mathds{P}_t: \mathcal{A}^n \rightarrow [0,1]$. 
For a fixed timestamp $t$, the goal of every agent $i$ is to determine a policy $\pi_i(A, s|\theta) \rightarrow [0, 1]$ with $A:=A(t)$ and $s:=s(t)$, that maximizes the long-term reward:
\begin{align}
  R_i = \smashoperator[r]{\sum_{t=0}^{\infty}} \gamma_t * r_i(t)
\end{align}
where $0 \leq \gamma_t \leq 1$ is a discount factor for time period $t$.

\subsection{Preprocessing and Model Architecture}
The RL agents are set up using a slight variation of the same baseline parametrization\footnote{cf. Raflin et al.,~\shortcite{raffinStableBaselines3ReliableReinforcement2021} to see tested implementations and baseline parametrization for several RL algorithms.} as we aim to keep our experiment as general as possible to ensure fungibility in different environments. In order to decrease the number of actions and thus simplify the action selection process, we decided to use a discrete action space. Compared to current literature\footnote{cf. Calvano et al.,~\shortcite{calvanoArtificialIntelligenceAlgorithmic2018} and Hettich~\shortcite{hettichAlgorithmicCollusionInsights2021} for other discrete (deep) q-learning action space implementations.}, however, actions are selected based on the price set in the last episode $p_i(t-1)$ for an agent $i$. In order to discretize the actions (i.e., the price) within the economic environment, we generate an evenly-spaced logarithmic distribution. The new price is calculated as follows:
\begin{equation}
  p_i(t) = \ln(1+e^{(p_i(t-1)+a_i(t-1))})
\end{equation}

For the sake of reducing complexity as well as computational cost we restrict the action space to a size of $\left| \mathcal{A} \right| = 7$ and the maximum adjustment of a price step to 2, resulting in the discrete action space $\mathcal{A} = [\text{-}2, \text{-}0.14, \text{-}0.01, 0, 0.01, 0.14, 2]$\footnote{We also test different action space sizes (i.e., 21, 51, and 71 actions). Please refer to the technical appendix at github.com/mschlechtinger/PriceOfAlgorithmicPricing.}. Hence, they will be able to keep the same price or evoke an increase or decrease within 3 gradients (small/ moderate/ big adjustment). We choose a Softplus activation in order to restrict the agents from setting prices below 0 while facilitating a derivation of the function (as opposed to ReLU). In order to counteract potential price rises due to the Softplus activation and to improve the overall robustness of the learning process, we restrict the randomizer to initialize the prices $P$ randomly from the interval $[0.5,1.5]$ each at the beginning of every episode. Although we track the agents' capital, they are able to accumulate debts without any consequences in the game, which yields more efficient learning. 

\section{Experiments}
We investigate collusion in two scenarios. Our baseline scenario \textit{Scenario A} depicts three agents that act based on decisions proposed by their given algorithm. In \textit{Scenario B} we manipulate each agent's state so that they are only able to observe their own prices as opposed to every price on the market. With this experimental setup, we can investigate whether the algorithms change their behavior when the experimental setup makes it more difficult to achieve collusion (i.e., in Scenario B). We apply Ray's RLlIB to profit from an open-source, industry-standard RL-algorithm implementations.\footnote{To ensure full reproducibility, we attached the code to this work at github.com/mschlechtinger/PriceOfAlgorithmicPricing.}

Each scenario is made up of several sub-scenarios, where we vary the number of agents (3, 5), the algorithm (PPO, DQN), and the biases $\mu= 0, 0.5, 1$. Every run is repeated 5 times to control for outliers, resulting in 90 runs overall (60 for Scenario A, 30 for Scenario B). A run comprises 10000 episodes with 365 steps each. We average the prices of every step to an episode price as well as a step profit (average profit of all steps in an episode) before averaging every run within the same setting. Finally, we apply locally-weighted scatterplot smoothing~\cite{roystonLowessSmoothing1992} to the averaged data. In addition to the agent data, the graphs incorporate the competitive benchmark (CB$\mid$ i.e., 1.01) as well as the monopolistic price (MP$\mid$ i.e., 1.50). On the other hand, the monopolistic profit (also abbreviated as MP) shown in the profit graphs is calculated by computing the profit a single seller would make by selling at the monopolistic price, divided by the number of sellers. If the agents converge to a price above the competitive benchmark (i.e., the Bertrand Equilibrium), economists classify this as a collusive market outcome. We assume that in practice, ECommerce companies would likely utilize similar pricing software due to the prevalence of a leading product. We thus focus on the interaction between similar agents. However, we challenge these settings by adjusting the neural network neurons, the algorithm's learning rate, the number of agents as well as their time of entry,  or the number of action gradients within the scope of an ablation study expounded upon in the technical appendix. 

Due to our modified way of action space discretization, we define convergence in a slightly different way than previous literature ~\cite{calvanoArtificialIntelligenceAlgorithmic2018}. A run converges if the rolling standard deviation of the mean agents' prices $\overline{\sigma \big(P(t)\big)}$ falls below a threshold of 0.01 for more than 100 episodes and this state lasts until the end of a run. If $\sigma \big(P(t)\big)$ reaches a value above the threshold of 0.01, we restart the counter. If it lasts below the threshold, we define this as \textit{episodes until convergence} $t_{EUC}$. There is a legal implication in the chosen definition of convergence phases, as during cartel cases, jurists generally attempt to identify phases during a run in which the risk of collusive behavior is especially high. Unlike human collusive behavior, which can mostly be traced back to certain collusive acts, algorithmic collusion is not induced by one single action, but by specific episodic phases. Our definition of convergence is one possible attempt to narrow down critical phases that qualify as collusion.

To generate a factor for measuring a degree of collusion in an economic sense, we need a consistent measure across all simulations. In line with Calvano et al.~\shortcite{calvanoArtificialIntelligenceAlgorithmic2018}, we calculate the average profit gain of all firms in a run $\Delta$, defined as
\begin{equation}
  \Delta = \frac{\overline{\Pi}-\Pi_\text{CB}}{\Pi_\text{MP}-\Pi_\text{CB}},
\end{equation}
where $\overline{\Pi}$ represents the average profit upon convergence of all firms (i.e., after passing every $t_{EUC}$), $\Pi_\text{CB}$ is the profit in the Bertrand-Nash static equilibrium (i.e., the competitive benchmark), and $\Pi_\text{MP}$ constitutes the profit in a state of perfect collusion (i.e., the monopoly price). Thus, $\Delta = 0$ corresponds to the competitive outcome and $\Delta = 1$ to the perfectly collusive outcome. It is important to mention that collusion in an economic sense does not imperatively indicate collusion in a legal sense, however, it can be an indicator of the latter.

\subsection{Scenario A: Competition}

Scenario A represents the main competitive setting of this research. Every agent operates based on the decisions computed by its own algorithms and their neural networks. In various sub-scenarios, we employed three and five agents, averaging data generated by PPO and DQN algorithms.

\begin{figure}[h]
  \centering
  \includegraphics[clip, trim=0cm 0.6cm 0.4cm 1.4cm, width=\linewidth]{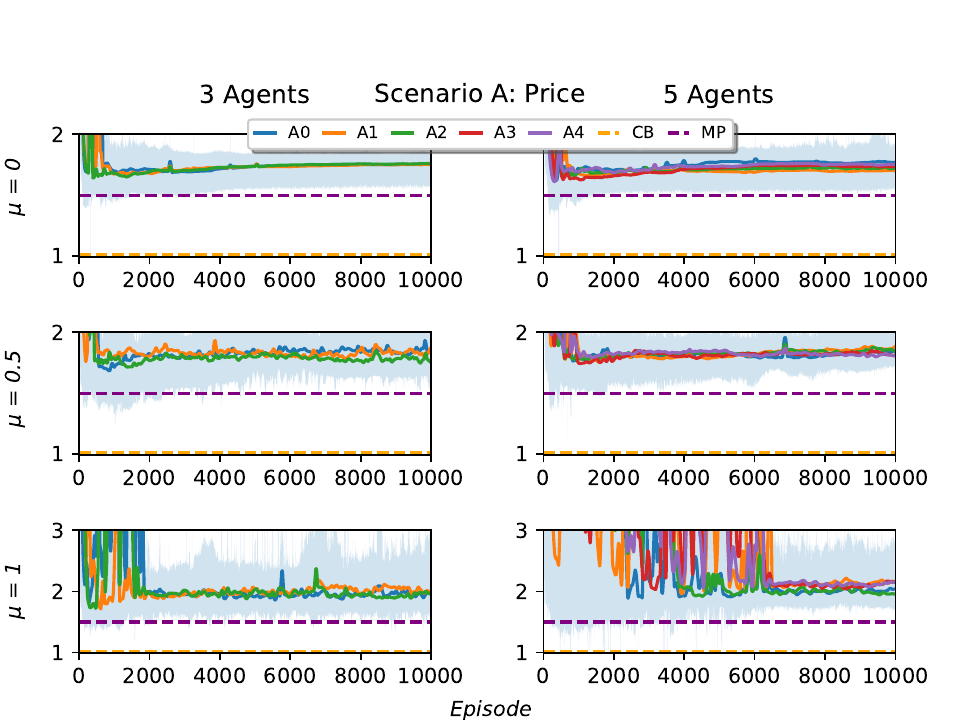}
  
  \includegraphics[clip, trim=0cm 0.6cm 0.4cm 0.0cm, width=\linewidth]{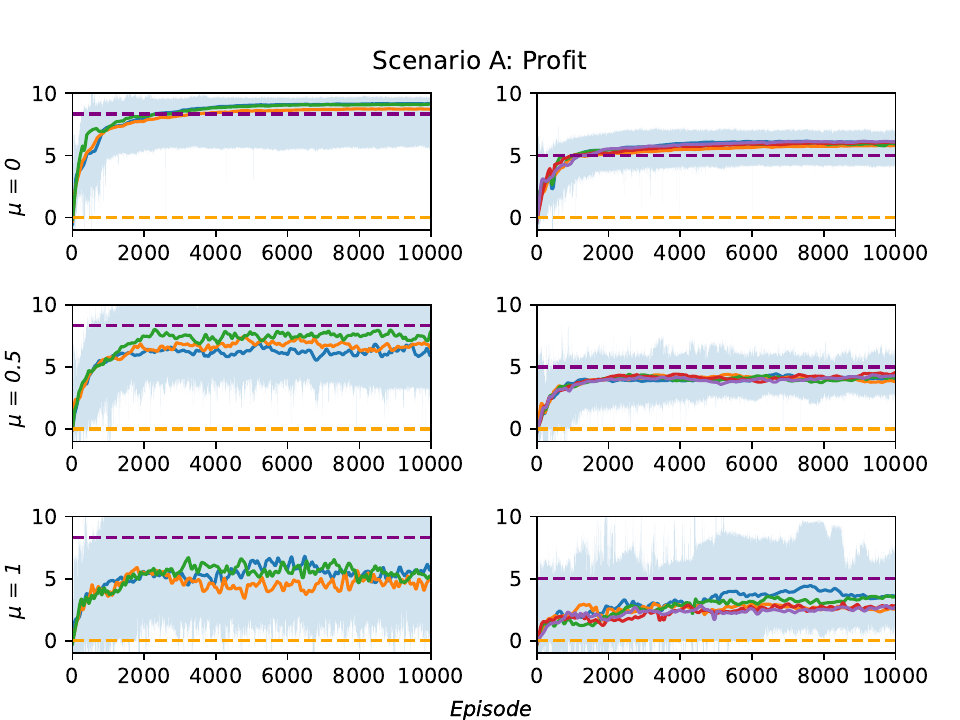}
  \caption{Episode Pricing and Profit in Scenario A}
  \label{fig:scenB_1}
\end{figure} 

Figure \ref{fig:scenB_1} presents the consolidated outcome of the runs through the price settings of three and five sellers on the left and right sides respectively. The pale blue area in each graph represents the variance of the runs. Although the competing agents exhibit distinct responses to the three bias weightings, each scenario leads to cooperative behavior that yields a supracompetitive price equilibrium in the long-term. These prices exceed the monopolistic price on average. However, the further we increase the price sensitivity ($\mu$) the higher the price-variance. We observe a similar, yet more extreme behavior when enhancing the number of sellers. The profit data (cf. Figure \ref{fig:scenB_1}) asserts these observations; the increased variance also significantly reduces earnings. 

\begin{figure}[h]
  \centering
  \includegraphics[clip, trim=0cm 0.6cm 0.4cm 0.8cm, width=\linewidth]{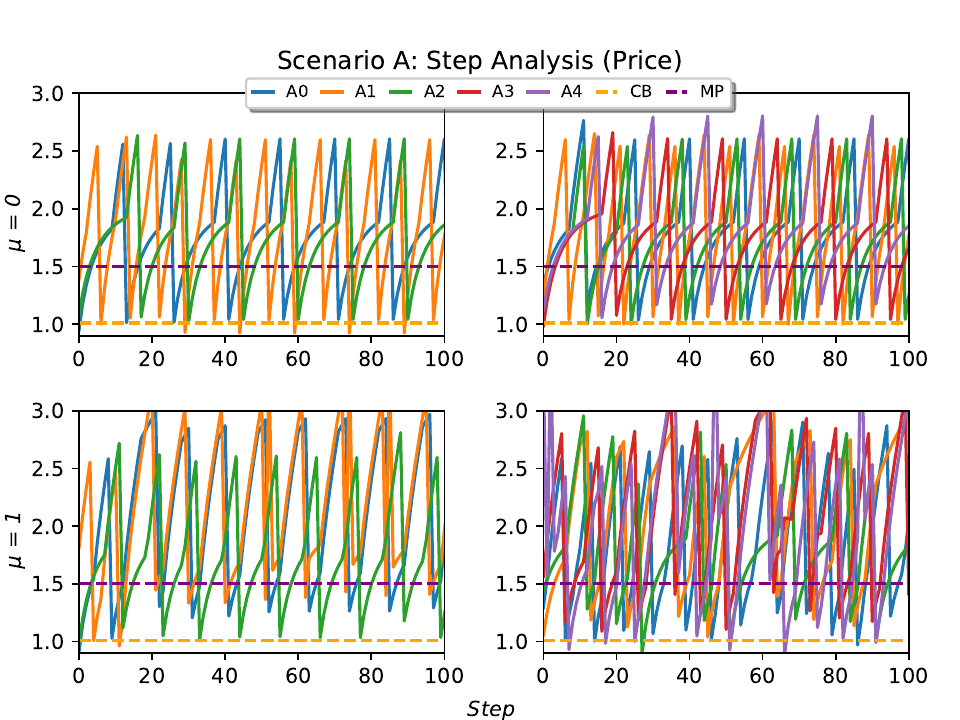}
  
  \includegraphics[clip, trim=0cm 0.6cm 0.4cm 0.6cm, width=\linewidth]{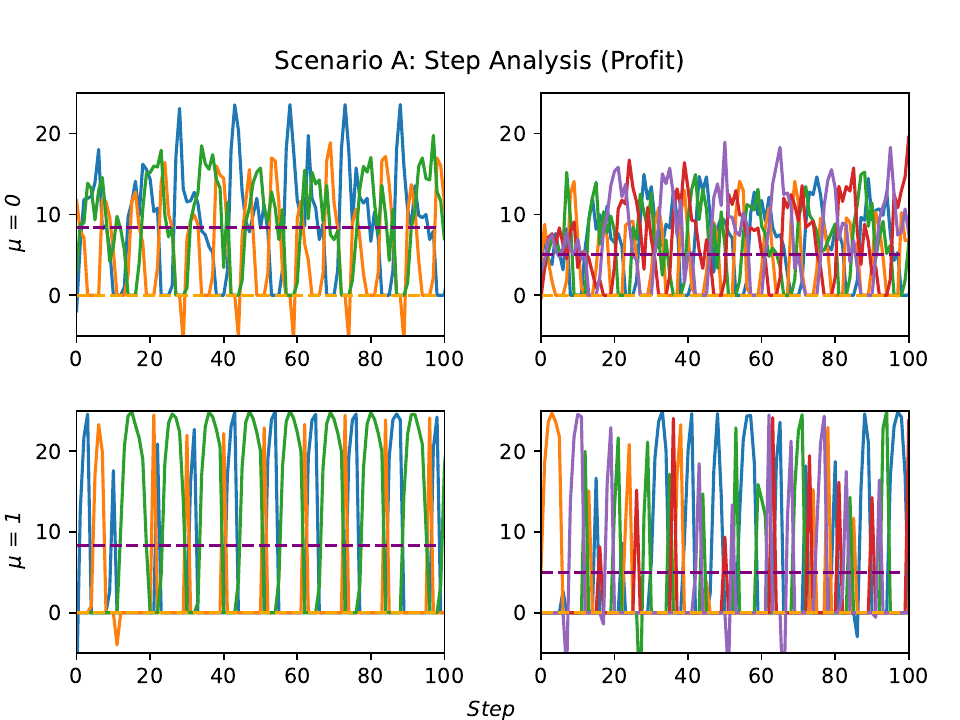}
  \caption{Step Pricing and Profit Excerpt in Scenario A}
  \label{fig:scenB_3}
\end{figure} 

To further scrutinize these results, we extracted the prices set and profits accumulated during the first 100 steps of the last episode in Figure \ref{fig:scenB_3}. Compared to the averaged episode overview, we can observe a price-setting behavior resembling an oscillation pattern, starting with a random price predefined by the environment. The agents occasionally slightly diverge from this cycle, however after a few steps, they usually get back in sync. These patterns explain why the prices exceed the monopolistic price on average; the agents find a strategy to collectively act as a monopolist on the market. To achieve this in a collaborative way, they traverse through three stages of pricing that are fundamentally the same but differ in execution in both $\mu=0$ and $\mu=1$. Initially, prices are set in close proximity to the competitive benchmark. Subsequently, prices are gradually escalated until a certain threshold is reached, beyond which they surpass the point of consumer demand saturation. Then, they return below the MP. The presented strategy exploits the boundaries set by the simulation by always having one agent earning the maximum. When $x=3$ and $\mu=0$ they abuse the lowered price sensitivity to create an increased demand with a higher price, thus maximizing the joint profit and exceeding the profit a monopolist would be able to achieve (i.e., $p=[1.01, 1.75, 1.75], \pi=[12.46, 0.66, 12.46]$). Similar patterns are discernible in instances where $\mu=1$. Notably, due to heightened price sensitivity, agents are unable to effectively stimulate increased demand. As a result, the agents endeavor to establish a monopolistic pricing regime, whereby two agents adopt prices above the threshold of $p=2.0$, while the third agent attempts to set a price near $1.50$ (i.e., $P=[2.0, 2.0, 1.50], \Pi=[0, 0, 25]$), with the objective of maximizing the joint profit. Although the oscillation patterns tend to remain consistent upon increasing the number of agents, identifying collaborative behaviors becomes more challenging when relying solely on visual inspection of the graphical representations.

\subsection{Scenario B: Constrained Observation Space}

In an attempt to weaken the ability to set supracompetitive prices, we constrain the agents' observation space in Scenario B to $s_i(t)=\{p_i(t), c\}$, thus removing the other competitors' prices from each agent's vision.

\begin{figure}[t]
  \centering
  \includegraphics[clip, trim=0cm 0.4cm 0.4cm 1.4cm, width=\linewidth]{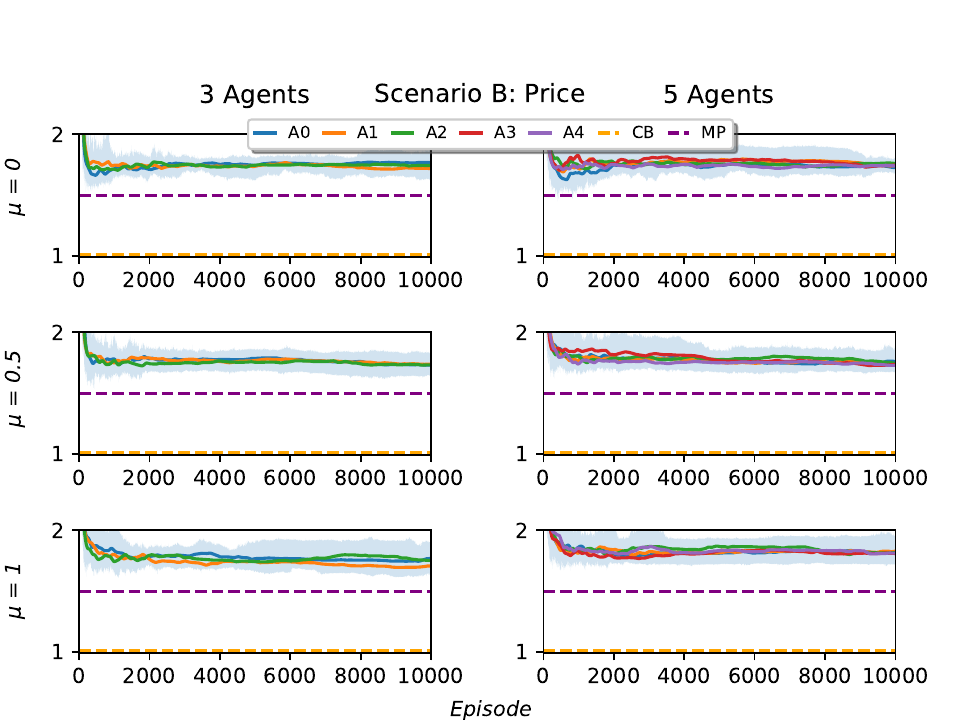}
  \centering
  \includegraphics[clip, trim=0cm 0.6cm 0.4cm 0.0cm, width=\linewidth]{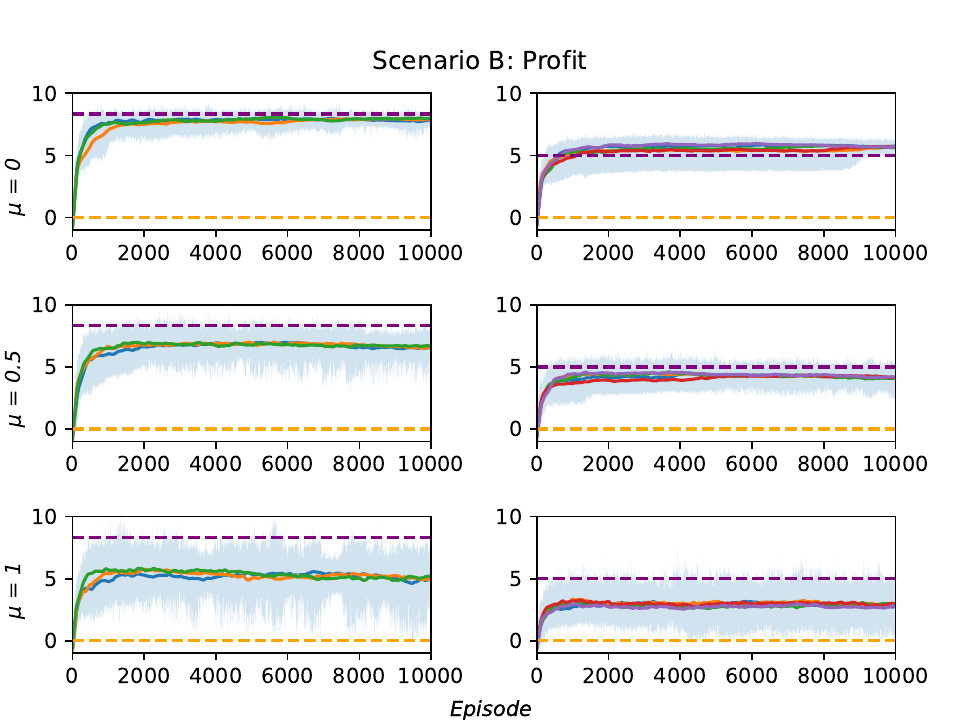}
  \caption{Episode Pricing and Profit in Scenario B}
  \label{fig:scenC}
\end{figure} 

The modification yields results that are comparable to those of Scenario A, under both run settings (3 and 5 agents) (cf. Figure \ref{fig:scenC}). In fact, contrary to our expectations, we observe less variance in the results. Analogously to Scenario A, we study a decrease in price with an increase of bias as well as an increase of variance when increasing the number of agents. The observations are reflected in the profit data.

\begin{figure}[h]
  \centering
  \includegraphics[clip, trim=0cm 0.4cm 0.4cm 0.8cm, width=\linewidth]{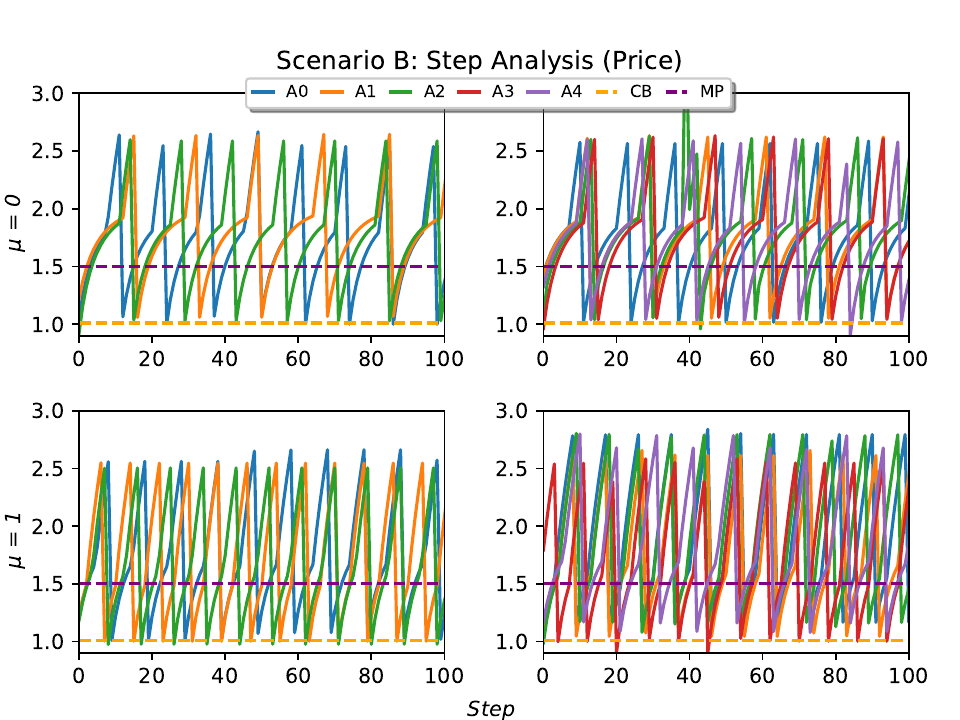}

  \includegraphics[clip, trim=0cm 0.6cm 0.4cm 0.6cm, width=\linewidth]{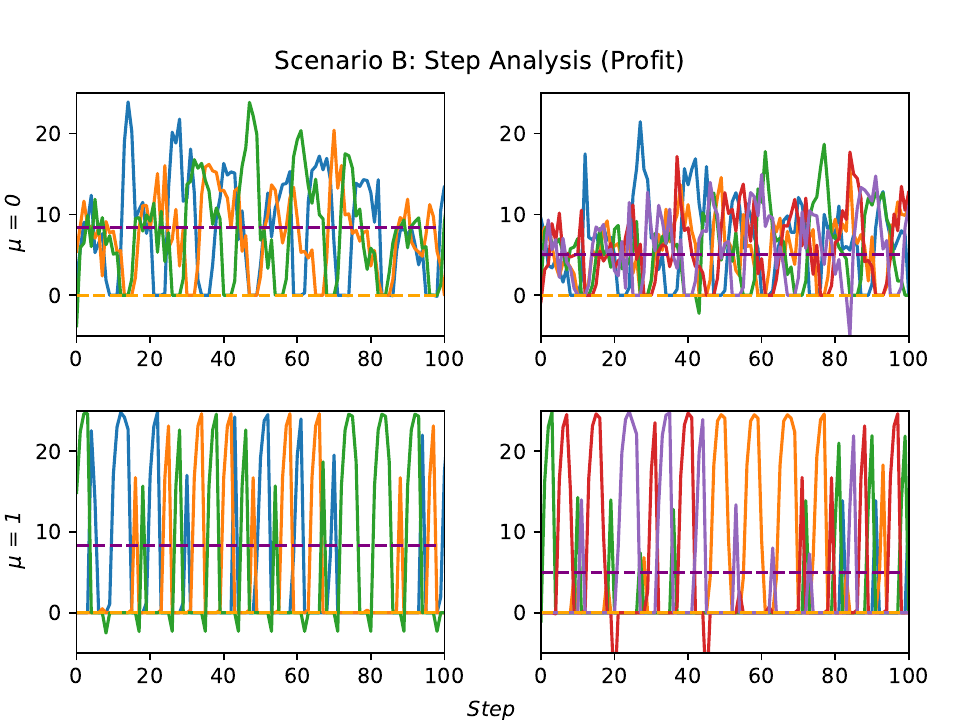}
  \caption{Step pricing and profit excerpt of Scenario B}
  \label{fig:scenC_3}
\end{figure} 

Upon investigating the step pricing (Figure \ref{fig:scenB_3}), we observe similar patterns to those in Scenario A. In both cases, where $\mu=0$ and $\mu=1$, the agents exhibit an oscillation pattern, resembling a slightly bent sawtooth shape. These observations are further confirmed by the step profit analysis, which depicts fluctuations between earning the minimum and earning above the monopolistic profit split.

\section{Discussion}
Our experiments have unveiled several noteworthy insights. First, our findings underscore the proficiency of DRL agents in establishing supracompetitive pricing strategies across a plethora of scenarios, without the need to disclose their competitors' pricing information. Second, our observations revealed the emergence of oscillation patterns within the agents' pricing behavior, which could be construed as an indicator of collaborative strategies. Third, the algorithms employed in our study yielded remarkable profitability, even more so when using PPO rather than DQN. This was exemplified by the agents achieving an average profit gain of $\Delta > 1$, indicating an outcome akin to perfect collusion, surpassing even monopolistic profit benchmarks. Fourth, we observed an overarching market stagnation that was achieved within a short timeframe and characterized by a consistently elevated $\Delta > 0.7$. Fifth, we underscore the robustness of our results in an ablation study (please refer to the supplementary material), revealing that modifications to fundamental simulation parameters fail to prevent a collusive outcome; conversely, they even enhance profit gains. These revelations collectively contribute to our understanding of the intricate dynamics of RL pricing agents in market scenarios and their inherent potential to attain collusive outcomes.


Although every run successfully implemented supracompetitive pricing strategies, subsequent analyses revealed that, within a theoretical Bertrand model, the agents encountered challenges in accumulating profits comparable to those observed in the unbiased runs. This can be attributed to multiple factors; the main reason for this weaker performance is the punishing nature of the theoretical scenario. Tying reward functions to the achieved profit immediately results in a punishment of exploration, which proves to be an important factor for collusive states (cf. Calvano et al.~\shortcite{calvanoArtificialIntelligenceAlgorithmic2018}). This finding is relevant to investigate how to prevent RL agents from colluding. Waltman and Kaymak~\shortcite{waltmanQlearningAgentsCournot2008} expressed that a force towards the collusive state is stronger if the agents get to experiment; if we can constrain the agents' ability to explore, we will thus experience less collusive behavior.

In line with Waltman~\shortcite{waltmanQlearningAgentsCournot2008}, we found that collusive states can be difficult to avoid in an oligopoly. This becomes particularly evident when examining the outcomes of the ablation study (please refer to the supplementary material). In contrast to the current literature, our outcome revealed that an increase in agents does not imply a decrease in prices.  Aligning with the current research trend of employing base-parametrized RL algorithms, we assert that optimizing these algorithms through proper tuning will only enhance the effectiveness of price-setting mechanisms in real-world scenarios.

One of the most notable findings from the data is the agents' capability to set prices above the competitive level without requiring access to their competitors' pricing information. Hansen et al.~\shortcite{hansenAlgorithmicCollusionSupracompetitive2020} already argued that the signal-to-noise ratio heavily affects results. We can only partly confirm those findings. While the average profit gains in Scenario B are slightly lower, we find that the overall outcome shares a striking resemblance to the results of Scenario A despite this severe confinement. The resiliency stems from the ability to approximate the prices via the reward function, in which other agents' prices embody unknown variables. This finding concurs well with Waltman~\shortcite{waltmanQlearningAgentsCournot2008}, who observed that agents without a memory were still able to collude.

In this context, it is important to mention that our step analysis experiments resemble the repricing patterns appearing in Amazon's Buybox~\cite{musolffAlgorithmicPricingFacilitates2022} as well as the observations by Klein~\shortcite{kleinAutonomousAlgorithmicCollusion2021}. We attribute two causes to the occurrence of these oscillating pricing patterns: First, the agents were able to increase short-term rewards with this strategy by increasing their profit above the monopolist's profit. Second, the discretization of the action space implies that the exact Bertrand and monopoly prices may not be feasible at all times, so the agents created mixed-strategy equilibria (cf. Calvano et al.~\shortcite{calvanoArtificialIntelligenceAlgorithmic2018} and the results of our ablation study).

While the results of our experiments show a collusive outcome in an economic sense, they do not undoubtedly exhibit whether algorithmic pricing constitutes permissible parallel behavior or a prohibited concerted practice and thus violates the cartel prohibition. Yet, they show that the agents seem to develop a degree of certainty about their competitors' anticipated next action to an extent that goes beyond mere observation of a single state and a logical adaption to it.

The cartel prohibition under Article 101 TFEU forbids any kind of joint conduct of independent market participants. According to the established case law of the European Court of Justice (ECJ), the characteristic of a concerted practice presupposes a minimum degree of coordination (concertation), a subsequent market conduct, and a causal link between the two.\footnote{cf. the cases~\cite{ecjCaseC199921999} para. 161;~\cite{ecjCaseC8082009} para. 38, 39;~\cite{ecjCaseC286132015} para. 125, 126.} This concertation does not have to be as binding as a contract or a direct agreement. It is sufficient that the uncertainty about the competitors' market behavior, which usually exists under competitive circumstances, is reduced.\footnote{cf. the cases~\cite{ecjCaseC8082009} para. 51;~\cite{ecjCase74142016} para. 39;~\cite{ecjCaseC286132015} para. 126.} However, there has to be at least an indirect contact between them because the cartel prohibition does not deprive them of the right to adapt their behavior to the observed or expected behavior of their competitors~\cite{ecjCaseC7951998}.

It is questionable whether these prerequisites, which follow human behavior's basic assumptions and logic, can equally be applied to RL decision-making processes. Based on our results, it could be argued that RL algorithms do not require any further reciprocity to gain the extra amount of trust in their competitors' expected next moves, which - in the case of human behavior - would be added through minimal contact. The insecurity about the competitors' next move is already reduced by the significant number of processed results from previous rounds, which are even indistinguishable from other environmental information and therefore inherent to each decision~\cite{schlechtingerWinningAnyCost2021}. 

It could be considered that the requirement of a minimum degree of communication might be obsolete because every important piece of information is explicitly or implicitly received via the reward function (cf. Scenario B). This raises the possibility that the reward function might channel competitor information, facilitating a concerted practice. However, the reward function simply symbolizes the agents' feedback on the profits achieved. Any market participant is allowed to know its own profit and draw conclusions about future pricing strategies from it. Given the aforementioned, the assumption that a collusive market outcome is inherently neutral if clear concertation is not detectable is debatable in the context of algorithmic pricing.

When conceptualizing simplified experiments, questions about the transferability of observations to reality and the validity of drawn conclusions quickly arise. We would like to emphasize that, while the present findings originate from a theoretical simulation, the overarching strategies hold the potential for broad applicability within real-world market contexts. As the economics of markets can be formulated as a (PO-)MDP, that - in theory - is solvable by RL algorithms, the agents will always strive for a policy that ultimately achieves the maximum reward. If this reward is tied to the profit, agents will realize that cooperation will help them achieve the best reward in the long run. 

As with every study, the results are beset with limitations which opens the door for future research. Although our experiments systematically investigated multiple algorithms, the exploration of potential interactions among heterogeneous algorithms was not pursued. While we present diverse scenarios in the context of an ablation study, we believe that further diversification of simulations will broaden our understanding of pricing algorithms. Moreover, future research endeavors could delve into strategies for imposing constraints on algorithms, thereby discouraging the emergence of collusive states. 

\section{Conclusion}
This paper utilizes an experimental approach to investigate algorithmic collusion. We find that deep RL agents using PPO and DQN are capable of charging supracompetitive prices without explicitly instructing them to do so. Furthermore, we have demonstrated that the algorithms will gravitate towards a collusive state, even when being restricted in their ability to conceive their competitors' prices. The results have undergone rigorous validation with the help of an ablation study.

\appendix



\section*{Acknowledgments}
The work presented in this paper has been conducted in the KarekoKI project, funded by the Baden-Wurttemberg Stiftung in the Responsible Artificial Intelligence program.

\bibliographystyle{named}
\bibliography{ijcai24}

\end{document}